\def \tool{\texttt{MIMONet}\xspace}

\documentclass[letterpaper, 10 pt, conference]{ieeeconf}  

\usepackage{booktabs}
\usepackage{amsfonts}
\usepackage{graphicx}
\usepackage[hidelinks]{hyperref}
\usepackage{textcomp}
\usepackage{listings}
\usepackage{adjustbox}
\usepackage{xspace}    
\usepackage{multirow}
\usepackage{multicol}
\usepackage{xcolor} 
\usepackage{tcolorbox}
\usepackage{wrapfig}
\usepackage{diagbox}
\usepackage{subcaption}
\usepackage{amsmath}
\usepackage{pifont}
\usepackage{breakcites}
\usepackage{array}
\usepackage{framed}
\definecolor{shadecolor}{rgb}{0.92,0.92,0.92}

\usepackage[linesnumbered,boxed,ruled]{algorithm2e}
\usepackage{listings}
\usepackage{bbding}
\let\labelindent\relax
\usepackage{enumitem}
\usepackage{subcaption}

\IEEEoverridecommandlockouts                              

\title{\LARGE 
\tool: Multi-Input Multi-Output On-Robot Deep Learning}

\author{Zexin Li$^{1}$, Xiaoxi He$^{2}$, Yufei Li$^{1}$, Wei Yang$^{3}$, Lothar Thiele$^{4}$, and Cong Liu$^{1}$%
\thanks{$^{1}$Authors are with the University of California, Riverside.}
\thanks{$^{2}$Authors are with the University of Macau.}
\thanks{$^{3}$Authors are with the University of Texas at Dallas.}
\thanks{$^{4}$Authors are with the ETH Zurich.}
}

\begin{document}

\renewcommand\arraystretch{1.1} 

\maketitle
\thispagestyle{plain}
\pagestyle{plain}

\begin{abstract}
Future intelligent robots are expected to process multiple inputs simultaneously (such as image and audio data) and generate multiple outputs accordingly (such as gender and emotion), similar to humans. Recent research has shown that multi-input single-output (MISO) deep neural networks (DNN) outperform traditional single-input single-output (SISO) models, representing a significant step towards this goal. In this paper, we propose \tool, a novel on-device multi-input multi-output (MIMO) DNN framework that achieves high accuracy and on-device efficiency in terms of critical performance metrics such as latency, energy, and memory usage. Leveraging existing SISO model compression techniques, \tool develops a new deep-compression method that is specifically tailored to MIMO models. This new method explores unique yet non-trivial properties of the MIMO model, resulting in boosted accuracy and on-device efficiency. Extensive experiments on three embedded platforms commonly used in robotic systems, as well as a case study using the TurtleBot3 robot, demonstrate that \tool achieves higher accuracy and superior on-device efficiency compared to state-of-the-art SISO and MISO models, as well as a baseline MIMO model we constructed. Our evaluation highlights the real-world applicability of \tool and its potential to significantly enhance the performance of intelligent robotic systems.
\end{abstract}
\section{INTRODUCTION}

Single-input single-output (SISO) deep neural networks (DNNs) have demonstrated impressive performance in various robotics applications~\cite{deng2020self, meng2019neural, yang2022instinctive, su2022uncertainty}. However, multi-input single-output (MISO) DNNs have emerged as a promising alternative, as they have been shown to surpass SISO DNNs both theoretically and empirically~\cite{DBLP:journals/corr/abs-2208-10442,DBLP:journals/corr/abs-2205-06175,DBLP:conf/iccv/HuS21}. Developing MISO DNNs is a crucial step towards creating multi-input multi-output (MIMO) DNNs for intelligent embedded systems, especially in the field of robotics.


According to Stanford’s ``Artificial Intelligence and Life in 2030'' report~\cite{stone2022artificial}, AI is expected to impact various fields, including home services~\cite{you2003development}, healthcare~\cite{ma2023learning,ma2022learning,lyu2022multimodal,lyu2024badclm}, and transportation~\cite{ma2023eliminating,ma2024data,cottam2024large,zhang2024large}. To build such robots thrive in these areas that inherently require MIMO systems, they must process real-time inputs like spoken language and facial expressions, providing personalized feedback based on factors like gender and emotion. Interactive companion robots, like the Vector Robot~\cite{bib:vector}, are expected to receive camera and microphone inputs and use this information to predict user preferences and respond appropriately. In on-device learning scenarios with strict hardware constraints and performance requirements, deploying multi-input multi-output (MIMO) deep neural networks (DNNs) can be advantageous. By reducing the need to deploy multiple single-input single-output (MISO) DNN instances on the device, MIMO DNNs can significantly lower resource demands. This paper presents \tool, a novel on-device MIMO DNN framework that achieves both high accuracy and on-device efficiency in terms of critical performance metrics including latency, energy, and memory usage. The MIMO approach inherently offers speed improvements by requiring only one forward pass for multiple tasks (instead of multiple passes), reducing computational redundancy.

We first constructed a baseline MIMO DNN model and observed that it requires high memory consumption and struggles to meet real-time constraints due to computational demand. To minimize resource demands, \tool builds on an existing DNN model compression technique, VIB~\cite{VIB}, which is effective for single-input single-output (SISO) DNN models. However, extending VIB to MIMO DNN models posed two new challenges. First, VIB exclusively supports the compression of feed-forward networks~\cite{Feed-Forward}, which makes it unsuitable for more advanced networks, such as ResNet~\cite{ResNet}. To address this limitation, \tool extends and develops a modified compression method for ResNet with residual blocks. Second, VIB was not designed to handle MIMO scenarios and did not account for the unique characteristics of MIMO models. VIB focuses on reducing intra-model redundancy but is unable to address common inter-model redundancy in MIMO models. To address this issue, \tool develops a new deep-compression method that reduces both inter- and intra-model redundancy, resulting in deeper lossless compression for MIMO scenarios.

We conducted a two-fold evaluation of \tool. First, we compared its accuracy against state-of-the-art SISO and MISO models using the RAVDESS dataset~\cite{RAVDESS}, showing that \tool outperforms them in most scenarios. In the second set, we conducted a comprehensive evaluation of \tool's components' impacts on on-device efficiency, including memory, latency, and energy. We deployed \tool on three widely used embedded system platforms: NVIDIA Jetson Nano Orin, NVIDIA AGX Xavier, and NVIDIA AGX Orin, which are most recent used platforms for various robotics applications~\cite{deng2020self, meng2019neural, yang2022instinctive,guo2023backdoor} such as Duckiebot~\cite{Duckiebot(DB-J)}, SparkFun Jetbot~\cite{SparkFun_JetBot}, and Waveshare Jetbot~\cite{Waveshare_JetBot}. In addition, we evaluated \tool on a PC machine for a more comprehensive assessment.
In summary, experimental results demonstrate that \tool can achieve:


\begin{itemize}[leftmargin=10px]
\item \textbf{Reduced Memory Usage:} \tool significantly reduces runtime memory usage by 80.7\% compared to the baseline MISO models.
\item \textbf{Improved Inference Speed:} \tool exhibits speed-ups of 1.98x, 2.29x, and 1.23x compared to the baseline MISO model when tested on Nano, AGX, and Orin, respectively.
\item \textbf{Enhanced Energy Efficiency:} The energy savings offered by \tool are 2.01x, 8.64x, and 2.71x compared to the baseline MISO model when tested on Nano, AGX, and Orin, respectively.
\end{itemize}


The major contributions of this paper can be summarized as follows:

\begin{itemize}[leftmargin=10px]
    \item \textbf{Innovative MIMO Framework for Robotics:} We introduce \tool, one of the first frameworks to utilize a unified neural network for processing multiple inputs and outputs simultaneously, addressing complex inference tasks in robotics.
    \item \textbf{New Deep-Compression Method:} \tool implements a novel deep-compression method that enhances accuracy and on-device efficiency, specifically tailored for MIMO applications in robotics.
    \item \textbf{Extensive Experimental Results:} Our experiments on three embedded platforms and a PC demonstrate \tool’s high accuracy and superior on-device efficiency in terms of latency, energy, and memory usage, making it suitable for robots under stringent hardware constraints.
    \item \textbf{Real-World Case Study Validation:} We conducted a realistic case study using the TurtleBot3 robot, showing \tool’s practical applicability and superiority over state-of-the-art SISO and MISO models in face and audio recognition within cluttered environments. This further validates its effectiveness for interactive robotic systems.
\end{itemize}

\section{BACKGROUND and RELATED WORK}
\subsection{MIMO Architecture}



Classical deep neural networks typically follow a single-input single-output (SISO) format, designed to process one input and produce one output at a time. However, as deep learning has evolved, certain networks have been adapted to handle multiple inputs from varied domains, enhancing accuracy. This approach, often referred to as multimodal learning~\cite{zhang2016multimodal} or multi-view learning~\cite{zhao2017multi}, leads to what is termed multi-input single-output (MISO) networks~\cite{DBLP:journals/corr/abs-2208-10442,DBLP:journals/corr/abs-2205-06175,DBLP:conf/iccv/HuS21,he2022towards}.
Building on this, models with advanced accuracies have progressed to multi-input-multi-output (MIMO) configurations~\cite{havasi2020training}, which are capable of generating several outputs in one go. This MIMO technology is particularly beneficial in fields like autonomous driving~\cite{liu2022bevfusion,huang2023fuller} and robotic navigation~\cite{chaplot2019embodied}, although it does come with substantial demands on memory and computational resources. Few concurrent works like BEVFusion~\cite{liu2022bevfusion}, try to alleviate these demands by optimizing data processing from sensors to mitigate performance issues.
Different from these works, our study aims to tackle these challenges by focusing on model compression to facilitate practical deployment in real-world scenarios.


\subsection{Model Compression for On-device Scenarios}

Modern DNNs, like the ViT-Huge~\cite{wu2020visual} and GPT-3~\cite{brown2020language}, although achieving state-of-the-art accuracy strain memory and computation resources for resource-constrained devices during inferences.
A practical solution to mitigate such an issue is model compression, which aims to reduce model parameter count without affecting accuracy. 
They can involve removing redundant parameters \cite{VIB,MTZ,bib:NIPS17:Dong,bib:NIPS93:Hassibi,bib:TMC21:Liu} or minimizing redundant precision \cite{bib:NIPS15:Courbariaux}.
Two subclasses of model compression for multi-branch models consist of single-model pruning and cross-model compression. The former compresses a single network by eliminating unnecessary operations and parameters, while the latter minimizes inter-model redundancy by identifying and removing shared parameters and operations across models \cite{VIB,MTZ,bib:NIPS17:Dong,bib:NIPS93:Hassibi,bib:CVPR19:Molchanov,bib:PIEEE20:Deng,bib:KDD21:He,bib:IJCAI18:Chou}. Notably, the Multi-Task Zipping (MTZ) approach has pioneered cross-model compression by automating the merging of pre-trained DNNs \cite{MTZ}.
However, \tool leverages the Variational Information Bottleneck (VIB) method to reduce intra-model redundancy, extending it for application to ResNet architecture. Furthermore, it adapts modified MTZ to reduce inter-model redundancy, which further
decrease the total parameters in the proposed MIMO model.
\section{METHODOLOGY}


\subsection{Overview of \tool}
\label{sec:overview}

\begin{figure*}[!hbt]
    \centering
    \begin{subfigure}[b]{0.45\textwidth}
        \centering
    \includegraphics[width=\textwidth]{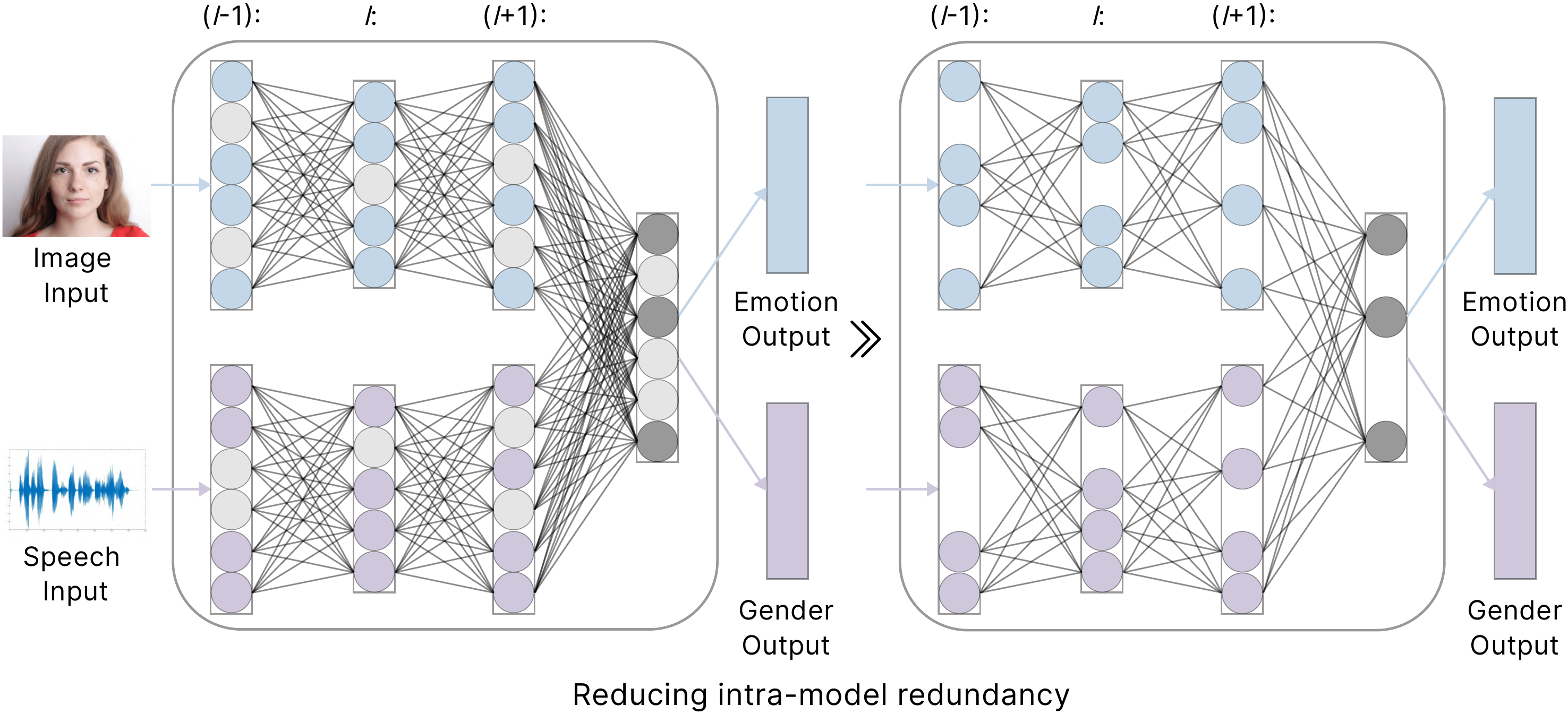}
    \label{fig:intra_model}
    \end{subfigure}
    \hfill
    \begin{subfigure}[b]{0.43\textwidth}
        \centering
    \includegraphics[width=\textwidth]{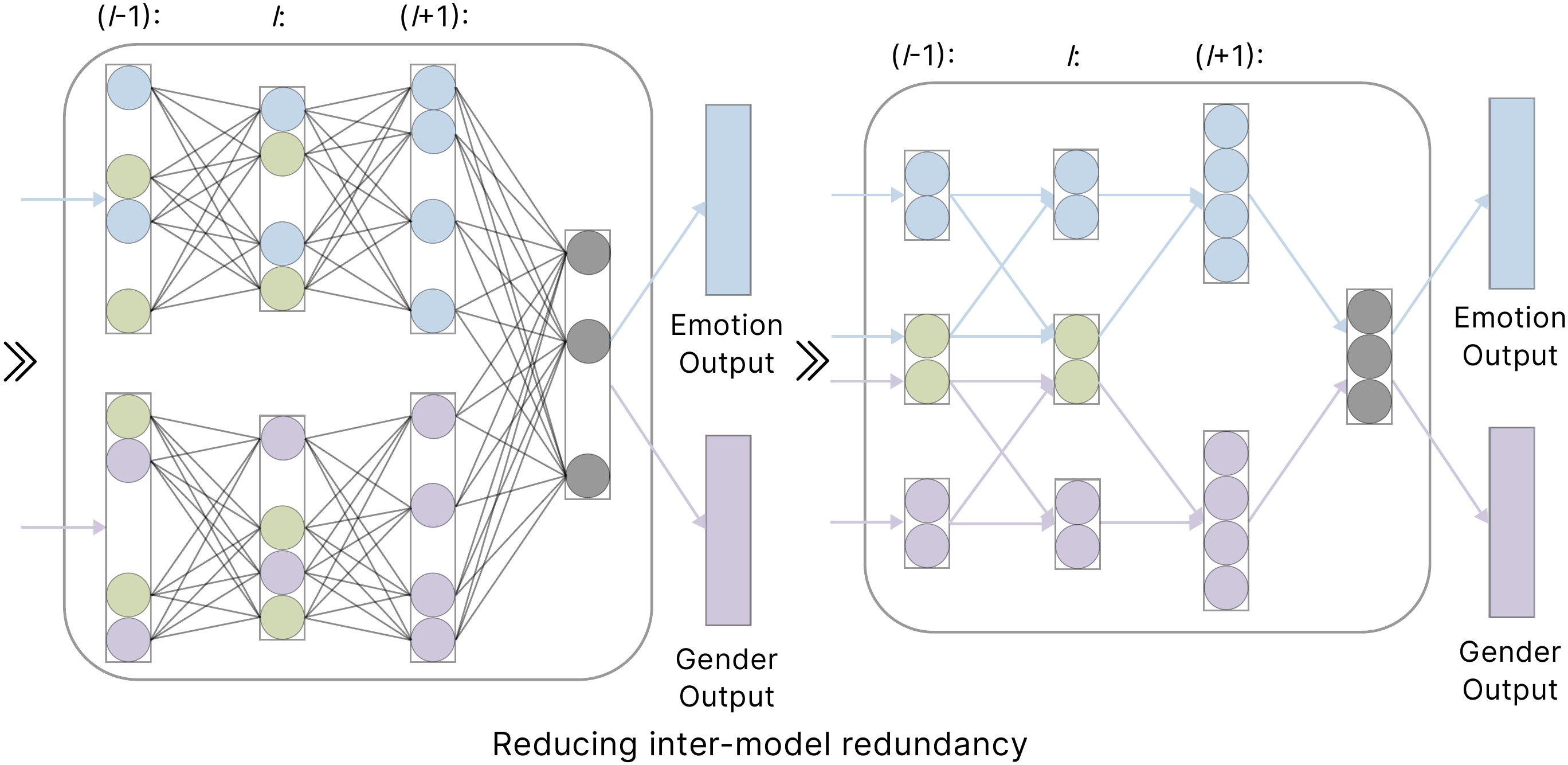}
    \label{fig:inter_model}
    \end{subfigure}
    \vspace{-5mm}
    \caption{Overview of \tool. In the left part, gray circles represent neurons inducing intra-model redundancy. In the right part, green circles denote sharable neurons inducing inter-model redundancy. Best viewed in color.}
    \label{fig:framework}
    \vspace{-5mm}
\end{figure*}


Fig.~\ref{fig:framework} illustrates an overview of the on-device MIMO inference framework. First, inspired by the famous multi-branch architecture ResNext~\cite{xie2017aggregated}, 
we adopt a MIMO deep neural network with multiple inputs backbone and multiple outputs classifiers as our baseline model (as shown in the first block). One forward prediction of this network requires multiple inputs (image and speech) and generates multiple prediction outputs (emotion and gender) simultaneously. To align and fuse feature information from the input image and speech, we use ResNet as the backbone to extract features from both modalities. The feature maps from both are then concatenated to create a unified representation. This fused feature map serves as the input for the subsequent prediction layers.

However, directly deploying such baseline MIMO models to embedded devices remains challenging. Specifically, embedded devices including most robots have \emph{limited memory} and require \emph{real-time performance} for many real-world scenarios. For instance, handling streaming data (e.g. high-speed cameras sampling 60 times per second) and ensuring robustness in safe-critical scenarios (e.g., autonomous driving robots~\cite{Duckiebot(DB-J),zhouhush,kong2019physgan,song2024bundledslam,song2024eta}). In addition, \emph{energy savings} are particularly needed for embedded devices because of limited battery capacity and heat dissipation.

To address these challenges, employing model compression is an intuitive and effective direction. First, we adopt the idea of VIB~\cite{VIB} to perform channel-wise hard masking (the masked channel generates zero output -- which can be pruned, while the unmasked channel's output remains the same) and follow pruning. Then, the pruned MIMO model can maintain the original structure in the forward direction.
However, VIB is designed for the SISO model and thus reduces the intra-model redundancy. However, it does not consider the characteristics of the MIMO framework, for instance, the MIMO framework is naturally multi-branching and exhibits inter-model redundancy. Therefore, aiming to reduce such redundancy intuitively can boost model compression performance. Furthermore, inspired by the idea from one cross-model compression work MTZ~\cite{MTZ}, we perform weights merging between multiple independent branches to improve model compression efficacy further. 
In summary, in \tool we focus on both intra-model redundancy and inter-model redundancy, and thus can theoretically further improve the compression effectiveness and achieve efficient on-device deployment.

\subsection{Reduce Intra-model Redundancy}

We aim to reduce intra-model redundancy by performing single-model compression. Specifically, we adopt and extend the idea from VIB~\cite{VIB} that reduces the redundant mutual information between adjacent layers within the model. The vanilla VIB only applies to neural networks such as VGG~\cite{VGG}. However, when dealing with neural networks that possess a non-linear architecture, such as ResNet~\cite{ResNet}, which forms the foundation of our MIMO model, the direct application of the conventional VIB approach is not feasible.

\begin{figure}[!hbtp]
    \vspace{-3mm}
    \centering
    \includegraphics[width=0.4\textwidth]{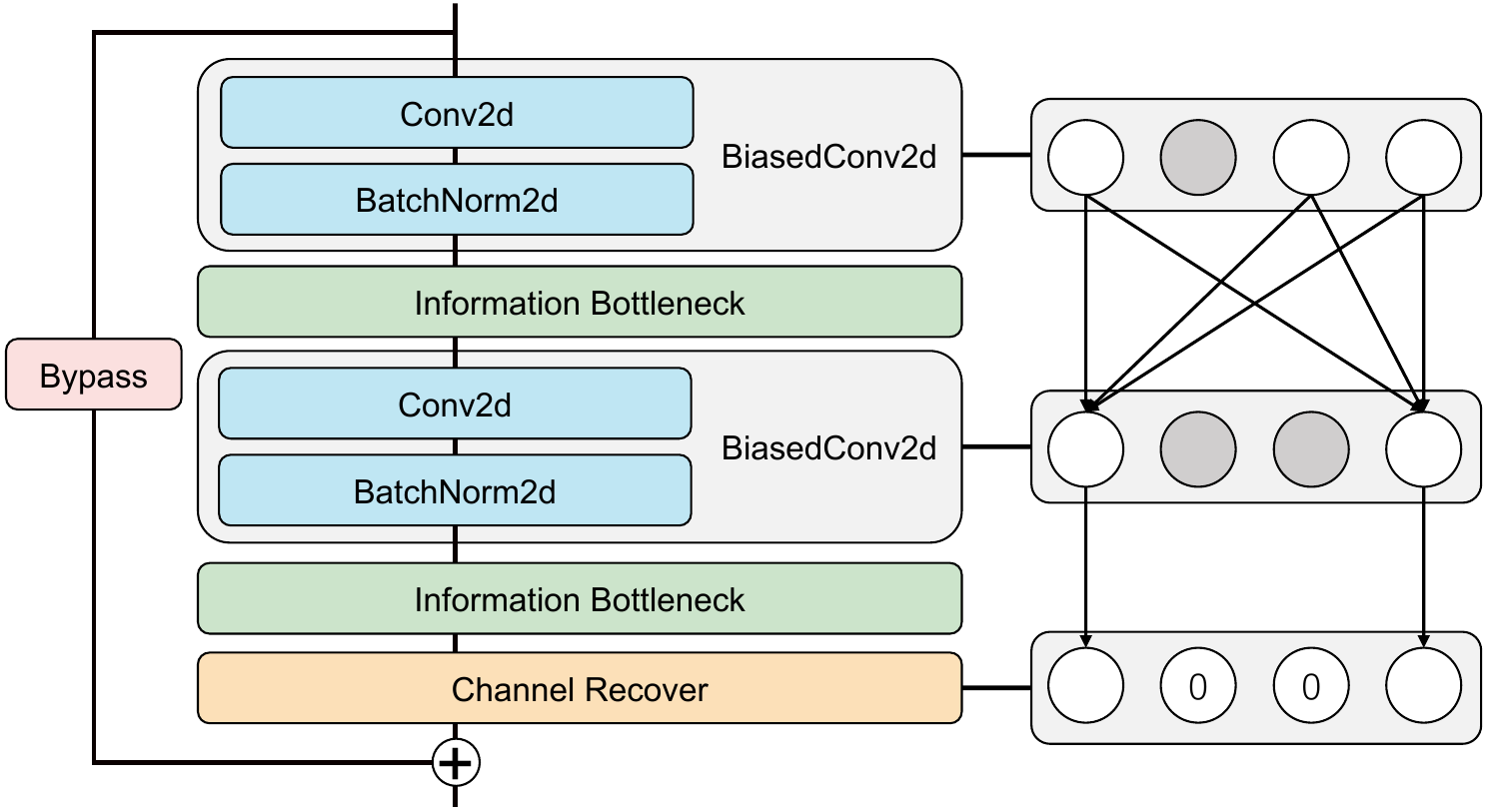}
    \caption{Design for compression of the residual block for ResNet~\cite{ResNet}. The left side shows the structure of the residual block. The right side shows channel-level pruning and recovery. White and gray circles exhibit kept pruned channels. The pruned channels are filled with zeros before the feature map summing. Best viewed in color.}
    \label{fig:residual_block}
    \vspace{-5mm}
\end{figure}

To address this, the design incorporates an ``information bottleneck" — a learnable mask — into each layer of the network that possesses learnable parameters (e.g., convolutional layers). This bottleneck selectively filters information flow between layers, allowing only the most relevant features to pass through, thus reducing redundancy.
As shown in the left part of Fig.~\ref{fig:framework}, neurons with gray color are masked. Furthermore, combining the characteristics of ResNet, we designed the following adaption scheme detailed in Fig.~\ref{fig:residual_block}: (1) Module-level design: we add the information bottleneck layer as in the original VIB design for the layers not in the residual block. However, we keep the input/output channel number unchanged for each residual block, i.e., we do not insert an information bottleneck layer between two residual blocks. (2) Block-level design: After the second information bottleneck layer, we insert a \textit{channel-recover} operation. This operation restores the channel to the pre-pruning mask based on the pruning mask by filling pruned channels with zeros, which ensures that the input channel is matched during the concatenation operation between the main branch and the bypass. Inspired by the design idea of ResNet~\cite{ResNet}, we intuitively assume the major information throughout the network is concentrated in the bypass (consisting of direct input or input after downsampling). Therefore, we do not set the mask in the bypass section.

The surprising result of this design is that according to top-5 layer-wise compression statistics upon application to the baseline MIMO model, the image branch is (100.0\%, 98.4\%, 97.9\%, 96.1\%, 87.1\%), and the speech branch is (100.0\%, 100.0\%, 99.4\%, 92.2\%, 86.7\%). The information bottleneck within the residual block naturally led to all output channels except the bypass from preceding layers being masked when a high compression rate was achieved, suggesting that the primary information indeed flows through the bypass. This outcome serves as empirical validation for the assumption underlying the adapted VIB approach for ResNet. The effectiveness of this method is further supported by the observed layer-wise compression rates, indicating the potential for significant model size reduction without compromising the essential information flow through the network.


Furthermore, we integrate performance optimization in the residual block to collapse each convolutional (Conv2d) layer and the subsequent batch normalization (BatchNorm2d) layer into a biased convolutional (BiasedConv2d) layer, as shown in the blue box in Fig.~\ref{fig:residual_block}. Since the on-device deployed models mainly only require forward inference, i.e., all parameters are fixed. Therefore, we could replace the BatchNormalization (BN) layer~\cite{ioffe2015batch} with multiplication and addition. Specifically, the output of the BN layer applied on the $i$-th channel of layer $l$ is:
\begin{equation}
    \vspace{-2mm}
	BN(y_{l,i}) =  \gamma_{l,i} \cdot \frac{y_{l,i}-\mu_{l,i}}{\sqrt{\sigma_{l,i}^2 + \epsilon}} + \beta_{l,i}
\end{equation}
where $y_{l,i}$ is the pre-activation output of the convolution layer, $\gamma_{l,i}$ and $\beta_{l,i}$ are the two learnable parameters (scaling and shifting) for the BN layer, $\mu_{l,i}$ and $\sigma_{l,i}$ are the pre-calculated mean and standard deviation. The effect of the BN layer can be replaced by multiplying the incoming weight of convolutional layer $\mathbf{w}_{l,i}$ by a scalar $\frac{\gamma_{l,i}}{\sigma_{l,i}}$ and adding $\beta_{l,i}-\frac{\gamma_{l,i}\cdot\mu_{l,i}}{\sigma_{l,i}}$ to the bias $b_{l,i}$.

\subsection{Reduce Inter-model Redundancy}


After reducing intra-model redundancy, we explore the possibility of decreasing the total number of parameters by reducing the inter-model redundancy through cross-model compression techniques. 
In a MIMO model, it is not obvious that sharable knowledge exists between the network's branched parts, as they are processing inputs from an entirely different domain, e.g., one for audio and the other for video input.
However, as shown in~\cite{MTZ2}, sharable knowledge still exists within models designed for different input domains, as long as their outputs are fused and serve the same output tasks.
Specifically, we have observed a significant amount of sharable knowledge can be found within the later layers, as features are becoming more abstract at this stage, and these abstract features are used for the same tasks. 

To this end, we extend the MTZ~\cite{MTZ,MTZ2} framework to merge model weights from multiple independent branches. 
MTZ is a framework that automatically and adaptively merges deep neural networks for cross-model compression via neuron sharing.
t the core of MTZ is the neural similarity metric, derived from a second-order analysis of the model parameters. It measures the similarity in function between two neurons' incoming weights, which is given by:
\begin{align}
    & d[\tilde{\mathbf{w}}^A_{l,i},\tilde{\mathbf{w}}^B_{l,j}] = \frac{1}{2}(\tilde{\mathbf{w}}^A_{l,i}-\tilde{\mathbf{w}}^B_{l,j})^\top \nonumber \\
    & \cdot
\left((\tilde{\mathbf{H}}^A_{l,i})^{-1}+(\tilde{\mathbf{H}}^B_{l,j})^{-1}\right)^{-1} \cdot(\tilde{\mathbf{w}}^A_{l,i}-\tilde{\mathbf{w}}^B_{l,j})
\end{align}
where $\tilde{\mathbf{w}}^A_{l,i},\tilde{\mathbf{w}}^B_{l,j}\in\mathbb{R}^{\tilde{N}_{l-1}}$ are the incoming weights, and $\tilde{\mathbf{H}}^A_{l,i}, \tilde{\mathbf{H}}^B_{l,j}$ are the associated layer-wise Hessians~\cite{MTZ,MTZ2}.

Motivated by this principle, we treat each branch as independent computing sub-graphs in our MIMO model.
As shown in the right part of Fig.~\ref{fig:framework}, neurons with green color are shared between different branches.
The branches are aligned from the very first layer such that they can be merged and optimized by MTZ layer after layer.
This design can also deal with branched layers with different widths, as long as they share the same underlying structure, such as convolutional layers or residual blocks.
As proved by experiments, reducing inter-model redundancy effectively reduces the baseline MIMO model size for more efficient on-device deployment.

\subsection{Integrate with Quantization Techniques}

To further enhance the on-device deployment efficiency of our MIMO inference framework, integrating quantization techniques presents a promising avenue. Quantization involves converting a model's floating-point weights and activations to lower precision formats, such as fixed-point integers, which significantly reduces the model size. This process is crucial for embedded devices with stringent memory and computational resource constraints.

Our proposed method, which already incorporates strategies for intra-model and inter-model redundancy reduction, can naturally extend to include quantization. In the quantization process, the weights and activations of the pruned and merged MIMO model are mapped from floating-point to fixed-point representation. This mapping can be achieved through various quantization schemes, such as uniform or non-uniform quantization, with or without re-training (quantization-aware training or post-training quantization). For instance, post-training quantization could be directly applied to the compressed model to quickly reduce its precision without the need for further training. Specifically, for a parameter \( w \), the quantization to 8-bit (int8) and 4-bit (int4) integers can be expressed as:

\[ w_{\text{int8}} = \text{round}\left( \frac{w - \text{min}(W)}{\text{max}(W) - \text{min}(W)} \times 255 \right) - 128 \]

\[ w_{\text{int4}} = \text{round}\left( \frac{w - \text{min}(W)}{\text{max}(W) - \text{min}(W)} \times 15 \right) - 8 \]

where \( W \) represents the set of all parameters in the model that are subject to quantization, \( \text{min}(W) \) and \( \text{max}(W) \) denote the minimum and maximum model parameters, respectively. This process introduces a trade-off between efficiency and accuracy. The integration with quantization techniques not only further reduces the memory footprint and computational requirements of the MIMO model but also leverages the hardware accelerators often found in embedded devices, which are optimized for low-precision arithmetic operations. Consequently, this integration can lead to significant improvements in inference speed and energy efficiency, making our MIMO inference framework even more suitable for real-time applications on resource-constrained devices.
\section{EVALUATION}

\subsection{Experimental Setup}

\begin{figure}[!b]
    \centering
    \includegraphics[width=0.47\textwidth]{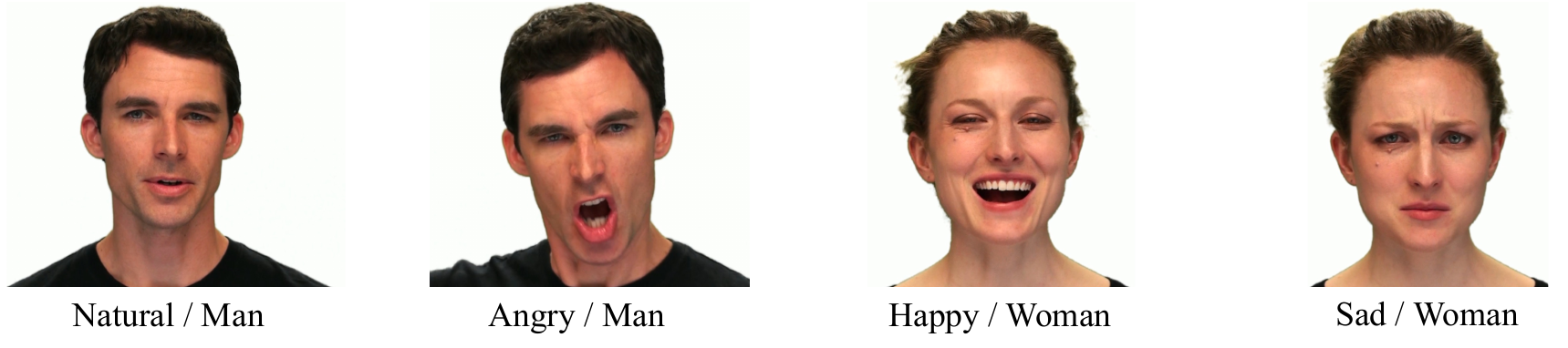}
    \caption{Data examples of RAVDESS dataset~\cite{RAVDESS}.}
    \label{fig:data_example}
    \vspace{-3mm}
\end{figure}

\begin{table*}[!t]
  \centering
  \caption{Experimental results of \tool. The best results are highlighted in bold. The second-best results are underlined.}
  \renewcommand\arraystretch{0.95}
  \label{tab:exp1}
  \resizebox{\textwidth}{!}{
    \begin{tabular}{l|cccc|ccc|ccc}
    \toprule
    \multirow{2}{*}{\textbf{Method}} & \multirow{2}{*}{\textbf{Acc./emo}} & \multirow{2}{*}{\textbf{Acc./gen}} & \multirow{2}{*}{$\textbf{\text{Par.}} (\times 10^{6})$} & \multirow{2}{*}{$\textbf{\text{Mem.}}$(MB)} & \multicolumn{3}{c|}{\textbf{\text{Latency}}(ms)} & \multicolumn{3}{c}{\textbf{\text{Energy}}(mJ)} \\
    \cline{6-11}
    & & & & &
    \textbf{Nano} & \textbf{AGX} & \textbf{Orin} & 
    \textbf{Nano} & \textbf{AGX} & \textbf{Orin} \\
    \midrule
    SISO/img-emo & 62.45 & N/A  & 11.82  & 1003.49 & 6.24 & 7.46 & 2.42 & 49.30 & 298.40 & 134.56  \\
    SISO/img-gen & N/A  & 77.40 & 11.82  & 1000.12 & 6.04  & 7.50 & 2.38 & 47.60  & 298.02 & 131.76 \\
    SISO/aud-emo & 61.81 & N/A  & 11.82  & 983.49 & 3.04 & 4.30 & 1.92 & 24.20  & 171.90 & 71.62  \\
    SISO/aud-gen & N/A  & \textbf{88.24} & 11.82  & 956.92 & 2.98  & 4.16 & 1.88 & 23.62  & 166.42 & 70.12  \\
    MISO/emo & 73.04 & N/A   & 25.51  & 1362.10 & 10.08 & 8.30 & 3.36 & 80.24  & 348.60 & 161.28  \\
    MISO/gen & N/A  & 77.43 & 25.51  & 1342.87 & 9.78  & 8.22 & 3.30 & 77.76  & 343.58 & 158.60  \\
    \midrule
    MIMO & 74.26 & 83.99 & 25.51  & 1364.96 & 5.26 & 4.15 & 2.95 & 39.98  & 175.29 & 81.90  \\
    \midrule
    \textbf{MIMONet (FP32)} & 76.28 & 83.17 & 0.92 & 910.11 & 5.21 & 3.96 & 2.84 & 40.88  & 45.93 & 66.17  \\
    \textbf{MIMONet (FP16)} & \underline{76.29} & 83.16 & 0.92 & 688.15  & 5.15 & 3.72 & 2.81 & 40.42  & 43.15 & 61.54 \\
    \textbf{MIMONet (INT8)} & \textbf{76.40} & 84.48 & \underline{0.92} & \underline{577.17} & \underline{5.11}  & \underline{3.66} & \underline{2.74} & \underline{39.96}  & \underline{41.34} & \underline{59.98} \\
    \textbf{MIMONet (INT4)} & 75.72 & \underline{87.01} & \textbf{0.92} & \textbf{521.68} & \textbf{5.02}  & \textbf{3.61} & \textbf{2.70} & \textbf{39.26}  & \textbf{40.07} & \textbf{59.01}  \\
    \bottomrule
    \end{tabular}
}
\vspace{-3mm}
\end{table*}%
We set up four baselines for comparison with \tool: two SISO models and two MISO models, both using ResNet-18 as the backbone. In MISO models, ResNet-18 extracts features from each input (image and audio), which are then concatenated and passed through three fully connected layers for prediction. \tool differs by having multiple predictors after the fully connected layers, enabling it to generate multiple predictions in a single forward pass. We also apply approximations like half-precision (FP16) and post-training quantization (INT8 and INT4).

We leverage the RAVDESS dataset~\cite{RAVDESS}, with 7356 multimodal clips from 24 actors, to evaluate our method’s performance in multi-input scenarios, specifically for concurrent emotion and gender predictions. Some examples are shown in Fig.~\ref{fig:data_example}. We use 80\% of the data (19 actors) for training and 20\% (the other 5 actors) for testing. All experiments are run three times, with averaged results reported.
Such tasks are pivotal for developing robots designed for human-robot interactions, applicable in healthcare for patient monitoring by interpreting emotions, in customer service for personalized assistance, and in education to tailor teaching strategies based on emotional and demographic insights.

\subsection{Metrics}

We study the following five metrics to evaluate the accuracy and on-device efficiency of \tool. 1) Accuracy: the ratio of correctly identified samples in the test set to the total number of samples in the test set. Note that the accuracies of the two tasks (emotion and gender) are evaluated separately. 2) Parameter number: number of learnable parameters. 3) Memory: required runtime memory for models. 4) Latency: average wall-clock time required to perform one forward prediction. 5) Energy: We measure system-wide energy consumption. For a fair comparison, we calculate the average latency and energy consumption for 10,000 forward passes. We aim to improve on-device efficiency (reduce parameter number, latency, memory, and energy) as much as possible while maintaining the model's effectiveness (accuracy).

\subsection{Effectiveness}
\label{sec:effectiveness}


\begin{table*}[!t]
  \centering
  \vspace{-1mm}
  \caption{Ablation study results of accuracy and on-device efficiency. The best results are in bold.} 
  \renewcommand\arraystretch{0.98}
  \label{tab:exp2}
  \resizebox{\textwidth}{!}{
    \begin{tabular}{l|cccc|ccc|ccc}
    \toprule
    \multirow{2}{*}{\textbf{Method}} & \multirow{2}{*}{\textbf{Acc./emo}} & \multirow{2}{*}{\textbf{Acc./gen}} & \multirow{2}{*}{$\textbf{\text{Par.}} (\times 10^{6})$} & \multirow{2}{*}{$\textbf{\text{Mem.}}$(MB)} & \multicolumn{3}{c|}{\textbf{\text{Latency}}(ms)} & \multicolumn{3}{c}{\textbf{\text{Energy}}(mJ)} \\
    \cline{6-11}
    & & & & &
    \textbf{Nano} & \textbf{AGX} & \textbf{Orin} & 
    \textbf{Nano} & \textbf{AGX} & \textbf{Orin} \\
    \midrule
    \textbf{MIMONet} & \textbf{76.40} & \textbf{84.48} & \textbf{0.92} & \textbf{577.17} & \textbf{5.11}  & \textbf{3.66} & \textbf{2.74} & \textbf{39.96}  & \textbf{41.34} & \textbf{59.98}  \\
    w/o VIB & 74.26 & 83.99   & 25.51 & 1364.96  & 5.26 & 4.15 & 2.95 & 39.98  & 175.29 & 81.90  \\
    w/o MTZ & 75.63 & 82.70 & 1.07  & 930.09 & 5.15 & 3.98 & 2.80 & 40.28 & 44.23  & 62.34 \\
    w/o PTQ & 76.28 & 83.17 & 0.92 & 910.11 & 5.32 & 3.96 & 2.84 & 40.88 & 45.93 & 66.17 \\
    \midrule
    \textbf{MIMONet} & \textbf{76.40} & \textbf{84.48} & \textbf{0.92} & \textbf{577.17} & \textbf{5.11}  & \textbf{3.66} & \textbf{2.74} & \textbf{39.96}  & \textbf{41.34} & \textbf{59.98}  \\
    w/ RandPruning & 12.66  & 51.22 & 0.92  & 579.22 & 5.15 & 3.71 & 2.77 & 40.43  & 42.23 & 60.12  \\
    \bottomrule
    \end{tabular}
}
\vspace{-6mm}
\end{table*}%

Table~\ref{tab:exp1} provides a comparative analysis of the \tool's performance against its SISO and MISO counterparts on the RAVDESS dataset. Observably, \tool and its variants with different precision (FP32, FP16, INT8, and INT4) consistently outperform SISO and MISO configurations across several metrics:

\begin{itemize}[leftmargin=*]
\item Accuracy: \tool models demonstrate superior performance, with the INT8 variant achieving the highest emotion recognition accuracy (Acc./emo) at 76.40\%, surpassing SISO and MISO models by 3.36\% and a relative gain of 4.6\%. The best gender recognition accuracy (Acc./gen) is 87.01\%, competitive with state-of-the-art results. Although SISO models excel in gender recognition using vocal features alone, adding image inputs can introduce variability, as facial cues are not always distinct. Nevertheless, \tool maintains excellent recognition ability even after compression due to ResNet’s efficient feature extraction and fusion of image and audio inputs. Compression techniques like INT8 further enhance efficiency by reducing resource requirements while preserving high accuracy, showcasing \tool’s adaptability in processing multi-modal data.

\item \textbf{Parameter number (Par.):} \tool largely reduce parameter number via reducing intra- and inter-task redundancies. It can reach a 96.4\% compression rate while achieving very competitive accuracy performance.
\item \textbf{Memory Usage (Mem.):} \tool demonstrates significant memory efficiency for memory-constrained robotic scenarios. For instance, while MISO/emo and MISO/gen configurations require about 1362.10 and 1342.87 MB of memory respectively, the \tool variants operate with markedly less, with the MIMO (INT4) requiring only 521.68 MB. Note that MIMO can operate both tasks simultaneously, unlike the two separate MISO models, which in total consume 2704.97 MB\footnote{The 2704.97 MB memory usage for two MISO models may double-count shared PyTorch framework memory. In practice, models can share some memory, reducing the total. However, \tool still provides substantial memory savings.}, representing approximately up to \textbf{80.7\%} reductions\footnote{Although the parameter count of \tool is reduced by up to 96.4\%, memory consumption is not reduced by as much. This is because the PyTorch framework reserves some internal memory and introduces memory overhead.} in memory requirements compared to MISO models.
\item \textbf{Latency (Latency/FP):} \tool shows better latency performance on all three embedded testbeds. Note that \tool could generate results of two tasks in one forward pass, while SISO or MISO requires two forward passes. Specifically, \tool exhibits speedup compared to MISO models up to 1.98x, 2.29x, and 1.23x on Nano, AGX, and Orin, respectively.
\item \textbf{Energy (Energy/FP):} \tool also demonstrates superior efficiency on all three embedded testbeds. Specifically, \tool exhibits energy saving compared to MISO models up to 2.01x, 8.64x, and 2.71x on Nano, AGX, and Orin, respectively.

\end{itemize}

\begin{minipage}{0.45\textwidth}
\begin{shaded}
    \textit{\textbf{Effective and Efficient on-device deployment}}: our experimental results demonstrate that \tool can address all challenges in Sec.~\ref{sec:overview} (limited memory, real-time constraints, and energy constraints) while not sacrificing and often improving accuracy performance.
\end{shaded}
\end{minipage}

\subsection{Ablation Study}
\label{sec:ablation}

We exhibit the ablation study results in Table~\ref{tab:exp2}. We interpret the results as follows:

\begin{itemize}[leftmargin=*]

\item \textbf{Efficacy of Reducing Intra-Task Redundancy (VIB):} Removing VIB increases memory usage from 577.17 MB to 1364.96 MB and parameters from 0.92 million to 25.51 million, highlighting its effectiveness in reducing intra-task redundancy. Unlike random pruning, which largely degrades accuracy, VIB retains relevant information and reduces noise, making it crucial and particularly work for managing MIMO tasks efficiently.

\item \textbf{Efficacy of Reducing Inter-Task Redundancy (MTZ):} Omitting MTZ reduces accuracy in emotion and gender recognition, increases memory usage, and raises latency slightly, demonstrating its importance in minimizing inter-task redundancy and maintaining accuracy.

\item \textbf{Efficacy of Post-Training Quantization (PTQ):} PTQ significantly reduces the model footprint through low-precision data structures, with minimal impact on accuracy. It decreases latency and energy consumption, enhancing post-training efficiency without major accuracy loss.

\end{itemize}

\begin{minipage}{0.45\textwidth}
\begin{shaded}
\textit{\textbf{Component Efficacy}}: The ablation study confirms the critical role of each \tool component in ensuring optimal on-device performance.
\end{shaded}
\end{minipage}



\subsection{Case study on TurtleBot3}
\label{sec:case}
We conduct a case study using the TurtleBot3 robot to evaluate the real-world applicability of \tool. Specifically, we replace the TurtleBot3's main control device with a Jetson Nano and install an RGB camera, USB audio card, and speakers. Subsequently, a native speaker was invited to converse with the robot using happy tones while facial videos were captured. The robot will make sounds and specific movements based on the predictions. We evaluate three models (MISO, vanilla MIMO, and \tool) and find that MISO generated wrong results and exhibited the longest response time due to two-pass inferences. In contrast, vanilla MIMO and \tool both yield accurate results, while \tool yields significantly shorter response times (3.00x speedup than vanilla MIMO on average), which aligns with our evaluation (Sec.~\ref{sec:effectiveness} and Sec.~\ref{sec:ablation}). 

\section{Discussion on Future Directions}

Recent research on model compression has primarily focused on simpler models, often neglecting complex architectures like ResNet~\cite{ResNet}, Transformer~\cite{DBLP:conf/nips/VaswaniSPUJGKP17}, and complex graph-based models~\cite{GCN,SAGE,gatconv,afarin2023commongraph,gao2023mega,dong2022deep,dong2022utility,dong2023graph}. Neural architecture search offers promising methods for creating efficient designs suitable for rapid inference~\cite{zhang2023accuracy}. Our goal is to adapt \tool for these advanced architectures, optimizing them for on-device use by addressing latency, energy, and memory limitations in models~\cite{DBLP:journals/corr/abs-2208-10442,DBLP:journals/corr/abs-2205-06175,gao2024dlora,han2024parameter}. While our approach improves performance on GPU-based devices, integrating FPGAs~\cite{canis2011legup,zhang2021stealing}, autonomous vechicles~\cite{dong2023deep}, and emerging VR technologies~\cite{slocum2023going} into AI-optimized embedded systems highlights a shift towards software-hardware co-design~\cite{zhang2021hardware} for heterogeneous computing, paving the way for broader on-device model deployment. In complex and diverse multi-task models, compression gains might diminish with less overlap between complex multiple tasks. We leave this to future work, where we plan to explore adaptive compression strategies tailored for complex multi-task scenarios.
\section{CONCLUSION}

This paper introduces \tool, one of the first on-device Multi-Input Multi-Output (MIMO) Deep Neural Network (DNN) frameworks targeting robotic applications. \tool achieves high accuracy and on-device efficiency in terms of critical performance metrics, such as latency, energy, and memory usage. Extensive experiments on three widely used embedded platforms demonstrate that \tool can provide high accuracy while enabling superior on-device efficiency, particularly when compared to state-of-the-art Single-Input Single-Output (SISO) and Multiple-Input Single-Output (MISO) models. Our research is an important first step toward building efficient MIMO DNN frameworks for practical on-device learning scenarios, particularly for robots that often face hardware and performance constraints.

\clearpage

\bibliographystyle{IEEEtran}
\bibliography{icra25}

\end{document}